\documentclass{article}
\usepackage{nips14submit_e,times}

\usepackage{times}
\usepackage{graphicx} % more modern
\usepackage{subfigure}
\usepackage{wrapfig}

\usepackage{natbib}

\usepackage{amsthm}

\usepackage{amsmath}

\usepackage{amsfonts}

\usepackage{amsmath}

\usepackage{algorithm}
\usepackage{algorithmic}

\usepackage{hyperref}

\title{Equilibrated adaptive learning rates for non-convex optimization}

\author{
Yann N. Dauphin$^{1}$, Harm de Vries$^{1}$, Yoshua Bengio\\
Universit\'{e} de Montr\'{e}al
}

\nipsfinalcopy % Uncomment for camera-ready version

\begin{document}

\maketitle

\footnotetext[1]{Indicates first authors}
\setcounter{footnote}{1}

\begin{abstract}
Parameter-specific adaptive learning rate methods are computationally efficient
ways to reduce the ill-conditioning problems encountered when training
large deep networks. Following recent work that strongly suggests that most of the
critical points encountered when training such networks are saddle points,
we find how considering the presence of negative eigenvalues of the Hessian
could help us design better suited adaptive learning rate schemes.
We show that the popular Jacobi preconditioner has undesirable behavior
in the presence of both positive and negative curvature, and present theoretical and
empirical evidence that the so-called equilibration preconditioner is comparatively
better suited to non-convex problems. We introduce a novel adaptive learning rate scheme,
called ESGD, based on the equilibration preconditioner.
Our experiments show that ESGD performs as well or better than RMSProp in terms of convergence speed, always clearly improving over plain stochastic gradient descent\footnote{An implementation is freely available at \url{https://gist.github.com/ynd/f1ce7133a03ec54d6eb9}}.
\end{abstract}

\section{Introduction}
One of the challenging aspects of deep learning is the optimization of the
training criterion over millions of parameters: the difficulty comes
from both the size of these neural networks and because the training objective
is non-convex in the parameters. Stochastic gradient descent (SGD) has
remained the method of choice for most practitioners of neural networks
since the 80's, in spite of a rich literature in numerical optimization.
Although it is well-known that first-order methods considerably slow down
when the objective function is ill-conditioned, it remains unclear how to
best exploit second-order structure when training deep networks.
Because of the large number of parameters, storing the full Hessian
(or other full-rank preconditioner) or even a low-rank approximation is not practical,
making parameter specific learning rates, i.e diagonal preconditioners, one of the viable alternatives. The questions is how to set the learning rate for SGD adaptively, both over time and for different parameters,
and several methods have been proposed~(see e.g. \cite{schaul2013unit} and references therein).

On the other hand, recent work~\citep{Dauphin-et-al-NIPS2014-small,Choromanska-et-al-AISTATS2015}
has brought theoretical
and empirical evidence suggesting that local minima are with high
probability not the main obstacle to optimizing large and deep neural
networks, contrary to what was previously believed: instead, saddle points
 are the most prevalent critical points on the optimization path
(except when we approach the value of the global minimum). These saddle
points can considerably slow down training, mostly because the objective
function tends to be flat in many directions and ill-conditioned in the
neighborhood of these saddle points. This raises the question: can we take
advantage of the saddle structure to design good and computationally
efficient preconditioners?

In this paper, we bring these threads together. We first study diagonal preconditioners for
saddle point problems, and find that the popular Jacobi preconditioner has unsuitable behavior in the presence of both positive and negative curvature. Instead, we propose to use the so-called equilibration preconditioner and provide new theoretical justifications for its use in Section \ref{sec:equi}. We provide specific arguments why equilibration is better suited to non-convex optimization problems than the Jacobi preconditioner and empirically demonstrate this for small neural networks in Section \ref{sec:absolute}. Using this new insight, we propose a new adaptive learning rate schedule for SGD, called ESGD, that is based on the equilibration preconditioner.
In Section~\ref{sec:setup} we evaluate the proposed method on two deep autoencoder benchmarks. The results, presented in Section~\ref{sec:results}, confirm that ESGD performs as well or better than RMSProp. In addition, we empirically find that the update direction of RMSProp is very similar to equilibrated update directions, which might explain its success in training deep neural networks.

\section{Preconditioning}\label{sec:preconditioning}
It is well-known that gradient descent makes slow progress
when the curvature of the loss function is very different in separate directions. The negative gradient will be mostly pointing in directions
of high curvature, and a small enough learning rate have to be chosen in order to avoid divergence in the largest positive curvature direction. As a consequence, the gradient step makes very little progress in small curvature directions, leading to the slow convergence often observed with first-order methods.

\begin{figure}[t]
\centering{
\subfigure[Original]{
 \includegraphics[width=0.3\textwidth]{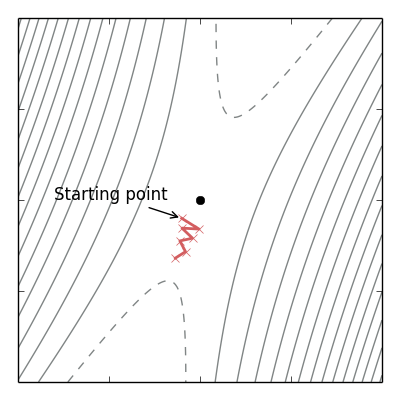}
}
\hspace{10mm}
\subfigure[Preconditioned]{
 \includegraphics[width=0.3\textwidth]{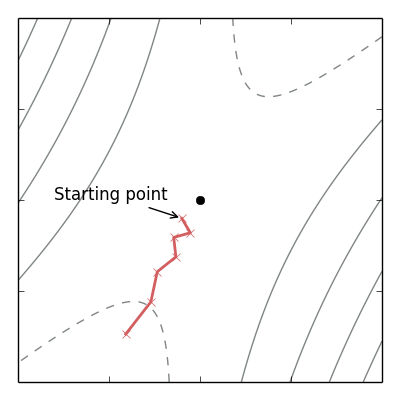}
}
\vspace{-1mm}
\caption{Contour lines of a saddle point (black point) problem for (a) original function and (b) transformed function (by equilibration preconditioner). Gradient descent slowly escapes the saddle point in (a) because it oscillates along the high positive curvature direction. For the better conditioned function (b) these oscillations are reduced, and gradient descent makes faster progress. }
\label{fig:contour-equilibration}
}
\vspace{-2mm}
\end{figure}

Preconditioning can be thought of as a geometric solution to the problem of pathological curvature. It aims to locally transform the optimization landscape so that its curvature is equal in all directions. This is illustrated in Figure~\ref{fig:contour-equilibration} for a two-dimensional saddle point problem using the equilibration
preconditioner (Section \ref{sec:equi}). Gradient descent method slowly escapes the saddle point due to the typical oscillations along the high positive curvature direction. By transforming the function to be more equally curved, it is possible for gradient descent to move much faster.

More formally, we are interested in minimizing a function $f$ with parameters $\bf{\theta} \in \mathbb{R}^N$. We introduce preconditioning by a linear change of variables $\hat{\bf \theta} = {\bf D}^{\frac{1}{2}} \mathbf{\theta}$ with a non-singular matrix ${\bf D}^{\frac{1}{2}}$. We use this change of variables to define a new function $\hat{f}$, parameterized by $\hat{\bf \theta}$, that is equivalent to the original function $f$:
\begin{equation}
 \hat{f}(\hat{{\bf \theta}}) = f({\bf D}^{-\frac{1}{2}}\hat{\bf \theta}) = f({\bf \theta})
\end{equation}
The gradient and the Hessian of this new function $\hat{f}$ are (by the chain rule):
\begin{align}
 \nabla \hat{f}(\hat{\theta}) &= {\bf D}^{-\frac{1}{2}} \nabla f({\bf \theta})\\
 \nabla^2 \hat{f}(\hat{\theta})&= {\bf D}^{-\frac{1}{2}\top} {\bf H} {\bf D}^{-\frac{1}{2}} \mbox{~~with~~} {\bf H} = \nabla^2 f({\bf \theta})
\end{align}
A gradient descent iteration $\hat{\bf \theta}_t = \hat{\bf \theta}_{t-1} - \eta \nabla \hat{f}(\hat{\theta})$ for the transformed function corresponds to
\begin{align}
 {\bf \theta}_t = {\bf \theta}_{t-1} - \eta {\bf D}^{-1} \nabla f({\bf \theta})
\end{align}
for the original parameter $\bf \theta$. In other words, by left-multiplying the original gradient with a positive definite matrix $D^{-1}$, we effectively apply gradient descent to the problem after a change of variables $\hat{\bf \theta} = {\bf D}^{\frac{1}{2}} \mathbf{\theta}$. The curvature of this transformed function is given by the Hessian ${\bf D}^{-\frac{1}{2}\top} {\bf H}{\bf D}^{-\frac{1}{2}}$, and we aim to seek a preconditioning matrix ${\bf D}$ such that the new Hessian has equal curvature in all directions. One way to assess the success of ${\bf D}$ in doing so is to compute the relative difference between the biggest and smallest curvature direction, which is measured by the condition number of the Hessian:
\begin{equation}
\kappa ({\bf H}) = \frac{\sigma_{\text{max}}({\bf H})}{\sigma_{\text{min}}({\bf H})}
\end{equation}
where $\sigma_{\text{max}}({\bf H}),\sigma_{\text{min}}({\bf H})$ denote
respectively the biggest and smallest singular values of ${\bf H}$ (which are the absolute value of the eigenvalues). It is important to stress that the
condition number is defined for both definite and indefinite matrices.

The famous Newton step corresponds to a change of variables ${\bf D}^{\frac{1}{2}} = {\bf H}^{\frac{1}{2}}$ which makes the new Hessian perfectly conditioned. However, a change of variables only exists\footnote{A real square root ${\bf H}^{\frac{1}{2}}$ only exists when ${\bf H}$ is positive semi-definite.} when the Hessian ${\bf H}$ is positive semi-definite. This is a problem for non-convex loss surfaces where the Hessian might be indefinite. In fact, recent studies~\citep{Dauphin-et-al-NIPS2014-small,Choromanska-et-al-AISTATS2015} has shown that saddle points are dominating the optimization landscape of deep neural networks, implying that the Hessian is most likely indefinite. In such a setting,  ${\bf H}^{-1}$ not a valid preconditioner and applying Newton's step without modification would make you move towards the saddle point. Nevertheless, it is important to realize that the concept of preconditioning extends to non-convex problems, and reducing ill-conditioning around saddle point will often speed up gradient descent.

At this point, it is natural to ask whether there exists a valid preconditioning matrix that always perfectly conditions the new Hessian? The answer is yes, and the corresponding preconditioning matrix is the inverse of the absolute Hessian
\begin{equation}
 |{\bf H}| = \sum_j |\lambda_j| {\bf q}_j {\bf q}_j^\top,
\end{equation}
which is obtained by an eigendecomposition of ${\bf H}$ and taking the absolute values of the eigenvalues. See Proposition 1 in Appendix A for a proof that $|{\bf H}|^{-1}$ is the only symmetric positive definite preconditioning matrix that perfectly reduces the condition number.

Practically, there are several computational drawbacks for using $|{\bf H}|^{-1}$ as a preconditioner. Neural networks typically have millions of parameters, rendering it infeasible to store the Hessian ($\mathcal{O}(N^2)$), perform an eigendecomposition ($\mathcal{O}(N^3)$) and invert the matrix ($\mathcal{O}(N^3)$). Except for the eigendecomposition, other full rank preconditioners are facing the same computational issues. We therefore look for more computationally affordable preconditioners while maintaining its efficiency in reducing the condition number of indefinite matrices. In this paper, we focus on diagonal preconditioners which can be stored, inverted and multiplied by a vector in linear time. When diagonal preconditioners are applied in an online optimization setting (i.e. in conjunction with SGD), they are often referred to as adaptive learning rates in the neural network literature.

\section{Related work}\label{sec:related}
The Jacobi preconditioner is one of the most well-known preconditioners. It is
given by the diagonal of the Hessian ${\bf D}^\text{J} = |\text{diag}({\bf H})|$
where $|\cdot|$ is element-wise absolute value. \cite{LeCun+98backprop} proposes an
efficient approximation of the Jacobi preconditioner using the Gauss-Newton
matrix. The Gauss-Newton has been shown to approximate the Hessian under certain
conditions \citep{Pascanu+Bengio-ICLR2014}. The merit of this approach is that
it is efficient but it is not clear what is lost by the Gauss-Newton
approximation. What's more the Jacobi preconditioner has not be found to be
competitive for indefinite matrices \citep{bradley2011matrix}. This will
be further explored for neural networks in Section \ref{sec:absolute}.

A recent revival of interest in adaptive learning rates has been started by AdaGrad
\citep{Duchi+al-2011}. Adagrad collects information from the gradients across
several parameter updates to tune the learning rate. This gives us the diagonal preconditioning matrix
${\bf D}^\text{A} = (\sum_t \nabla f^2_{(t)})^{-1/2}$
which relies on the sum of gradients $\nabla f_{(t)}$ at each timestep $t$.
\cite{Duchi+al-2011} relies strongly on convexity to justify this method. This
makes the application to neural networks difficult from a theoretical perspective. RMSProp \citep{tieleman2012lecture} and
AdaDelta \citep{Zeiler-2012} were follow-up methods introduced to be practical adaptive learning methods to train large neural networks. Although RMSProp has been shown to work very well~\citep{schaul2013unit}, there is not much understanding for its success in practice. Preconditioning might be a good framework to get a better understanding of such adaptive learning rate methods.

\section{Equilibration}\label{sec:equi}
Equilibration is a preconditioning technique developed in the numerical mathematics community~\citep{sluis1969condition}. When solving a linear system ${\bf A}x=b$ with Gaussian Elimination, significant round-off errors can be introduced when small numbers are added to big numbers~\citep{Datta:2010:NLA:1805893}. To circumvent this issue, it is advised to properly scale the rows of the matrix before starting the elimination process. This step is often referred to as row equilibration, which formally scales the rows of ${\bf A}$ to unit magnitude in some $p$-norm. Throughout the following we consider $2$-norm. Row equilibration is equivalent to multiplying ${\bf A}$ from the left by the matrix ${\bf D}^{-1} = \frac{1}{\|{\bf A}_{i, \cdot}\|}_2$. Instead of solving the original system, we now solve the equivalent left preconditioned system $\hat{\bf A}x=\hat{b}$ with $\hat{\bf A}={\bf D}^{-1}{\bf A}$ and $\hat{b}={\bf D}^{-1}b$.

In this paper, we apply the equilibration preconditioner in the context of large scale non-convex optimization. However, it is not straightforward how to apply the preconditioner. By choosing the preconditioning matrix
\begin{equation}
{\bf D}^\text{E} = \|H_{i, \cdot}\|_2,
\end{equation}
the Hessian of the transformed function $({\bf D}^\text{E})^{-\frac{1}{2}\top} {\bf H} ({\bf D}^\text{E})^{-\frac{1}{2}}$ (see Section \ref{sec:preconditioning}) does not have equilibrated rows. Nevertheless, its spectrum (i.e. eigenvalues) is equal to the spectrum of the row equilibrated Hessian $({\bf D}^\text{E})^{-1} {\bf H}$ and column equilibrated Hessian ${\bf H} ({\bf D}^\text{E})^{-1}$. Consequently, if row equilibration succesfully reduces the condition number, then the condition number of the transformed Hessian $({\bf D}^\text{E})^{-\frac{1}{2}\top} {\bf H} ({\bf D}^\text{E})^{-\frac{1}{2}}$ will be reduced by the same amount. The proof is given by Proposition 2.

From the above observation, it seems more natural to seek for a diagonal preconditioning matrix ${\bf D}$ such that ${\bf D}^{-\frac{1}{2}}{\bf H}{\bf D}^{-\frac{1}{2}}$
is row and column equilibrated. In \citet{bradley2011matrix} an iterative stochastic procedure is proposed for finding such matrix. However, we did not find it to work very well in an online optimization setting, and therefore stick to the original equilibration matrix ${\bf D}^\text{E}$.

Although the original motivation for row equilibration is to prevent round-off errors, our interest is in how well it is able to reduce the condition number. Intuitively, ill-conditioning can be a result of matrix elements that are of completely different order. Scaling the rows to have equal norm could therefore significantly reduce the condition number. Although we are not aware of any proofs that row equilibration improves the condition number, there are theoretical results that motivates its use. In \citet{sluis1969condition} it is shown that the condition number of a row equilibrated matrix is at most a factor $\sqrt{N}$ worse than the diagonal preconditioning matrix that optimally reduces the condition number. Note that the bound grows sublinear in the dimension of the matrix, and can be quite loose for the extremely large matrices we consider. In this paper, we provide an alternative justification using the following upper bound on the condition number from \citet{guggenheimer1995}:
\begin{equation}
 \kappa ({\bf H}) < \frac{2}{|\text{det } \bf H|} \left(\frac{\|\mathbf{H}\|_F}{\sqrt{N}}\right)^N
\end{equation}
The proof in \citet{guggenheimer1995} provides useful insight when we expect a tight upper bound to be tight: if all singular values, except for the smallest, are roughly equal.

We prove by Proposition 4 that row equilibration improves this upper bound by a factor $\text{det}({\bf D}^E) \left(\frac{\|\mathbf{H}\|_F}{\sqrt{N}}\right)^N$. It is easy see that the bound is more reduced when the norms of the rows are more varied. Note that the proof can be easily extended to column equilibration, and row and column equilibration. In contrast, we can not prove that the Jacobi preconditioner improves the upper bound, which provides another justification for using the equilibration preconditioner.

A deterministic implementation to calculate the $2$-norm of all matrix rows needs to access all matrix elements. This is prohibitive for very large Hessian's that can not even be stored. We therefore resort to a matrix-free estimator of the equilibration matrix that only uses matrix vector multiplications of the form
$({\bf H}{\bf v})^2$ where the square is element-wise and ${\bf v}_i \sim \mathcal{N}(0, 1)$\footnote{Any random variable ${\bf v}_i$ with zero mean and unit variance can be used. }. As shown by \citet{bradley2011matrix}, this estimator is unbiased, i.e.
\begin{align}
\label{eqn:equilibration}
\|{\bf H}_{i, \cdot}\|^2 = \mathrm{E}[({\bf H}{\bf v})^2].
\end{align}
Since multiplying the Hessian by a vector can be efficiently done without ever
computing the Hessian, this method can be efficiently used in the context of neural
networks using the R-operator \cite{Schraudolph02}. The R-operator computation
only uses gradient-like computations and costs about the same as two backpropagations.

\section{Equilibrated learning rates are well suited to non-convex problems}\label{sec:absolute}
\begin{figure}[t]
\begin{center}
\subfigure[convex]{\includegraphics[width=0.45\linewidth]{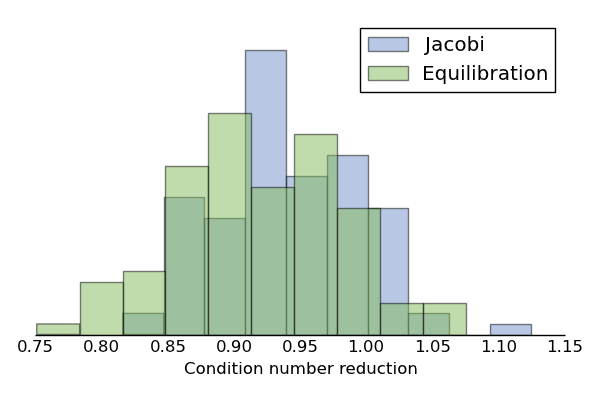}}
\subfigure[non-convex]{\includegraphics[width=0.45\linewidth]{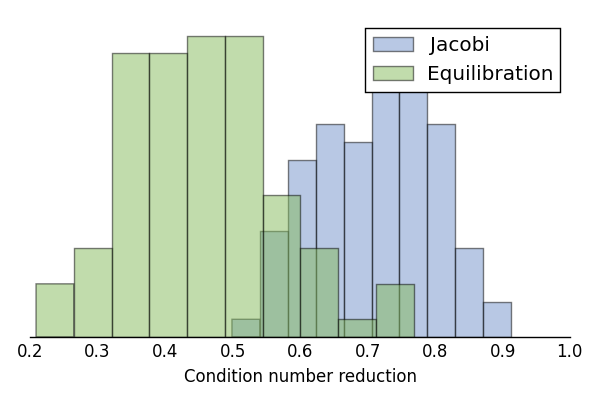}}
\vskip -0.1in
\caption{Histogram of the condition number reduction (lower is better) for random Hessians in a (a) convex and b) non-convex setting. Equilibration clearly outperforms the other methods in the non-convex case.}
\label{fig:reduction}
\vskip -0.1in
\end{center}
\end{figure}

In this section, we demonstrate that equilibrated learning rates are
well suited to non-convex optimization, particularly compared to the Jacobi
preconditioner. First, the diagonal equilibration matrix can be seen as an approximation
to diagonal of the absolute Hessian. Reformulating the equilibration matrix as
\begin{equation}
 {\bf D}^{\text{E}} = \|{\bf H}_{i, \cdot}\| = \sqrt{\text{diag}({\bf H}^2)}
\end{equation}
reveals an interesting connection. Changing the order of the square root and diagonal would give us the diagonal of $|{\bf H}|$.
In other words, the equilibration preconditioner can be thought of as the Jacobi preconditioner of the absolute Hessian.

Recall that the inverse of the absolute Hessian $|{\bf H}|^{-1}$ is the only symmetric positive definite matrix
that reduces the condition number to $1$ (the proof of which can be be found in Proposition 1 in the Appendix). It can be considered as
the gold standard, if we do not take computational costs into account. For indefinite matrices, the diagonal of the Hessian ${\bf H}$ and the diagonal of the absolute Hessian $|{\bf H}|$ will be very different, and therefore the behavior of the Jacobi and equilibration preconditioner will also be very different.

In fact, we argue that the Jacobi preconditioner can cause divergence because it
underestimates curvature. We can measure the amount of curvature in a
given direction with the Raleigh quotient
\begin{align}
R({\bf H}, {\bf v}) = \frac{{\bf v}^T{\bf H}{\bf v}}{{\bf v}^T{\bf v}}.
\end{align}
This quotient is large when there is a lot of curvature in the direction
${\bf v}$. The Raleigh quotient can be decomposed into
$R({\bf H}, {\bf v}) = \sum_j^N \lambda_j {\bf q}_j {\bf q}_j^T {\bf v}$ where
$\lambda_j$ and ${\bf q}_j$ are the eigenvalues and eigenvectors of ${\bf H}$.
It is easy to show that each element of the Jacobi matrix is given by
${\bf D}^\text{J}_i=|R({\bf H}, {\bf I}_{\cdot, i})|^{-1} = |\sum_j^N \lambda_j {\bf q}_{j,i}^2|^{-1}$.
An element ${\bf D}^\text{J}_i$ is the inverse of the sum of the eigenvalues
$\lambda_j$. Negative eigenvalues will reduce the total sum and make the step much
larger than it should. Specifically, imagine a diagonal element where
there are large positive and negative curvature eigendirections. The contributions
of these directions will cancel each other and a large step will be taken in that direction.
However, the function will probably also change fast in that direction (because of the high curvature),
and the step is too large for the local quadratic approximation we have considered.

Equilibration methods never diverge this way because they will not underestimate
curvature. In equilibration, the curvature information
is given by the Raleigh quotient of the squared Hessian
${\bf D}^\text{E} = (R({\bf H}^2, {\bf I}_{\cdot, i}))^{-1/2}=(\sum_j \lambda_j^2 {\bf q}_{j,i}^2)^{-1/2}$.
Note that all the elements are positive and so will not cancel. Jensen's
inequality then gives us an upper bound
\begin{equation}
{\bf D}^\text{E}_i \leq |{\bf H}|^{-1}_{ii}.
\end{equation}
which ensures that equilibrated adaptive learning rate will in fact be more conservative than the
Jacobi preconditioner of the absolute Hessian (see Proposition 2 for proof).

This strengthens the links between equilibration and the
absolute Hessian and may explain why equilibration has been found to
work well for indefinite matrices \cite{bradley2011matrix}. We have
verified this claim experimentally for random neural
networks. The neural networks have 1 hidden layer of a 100 sigmoid units with
zero mean unit-variance Gaussian distributed inputs, weights and biases. The
output layer is a softmax with the target generated randomly. We also give
results for similarly sampled logistic regressions. We compare reductions
of the condition number between the methods. Figure \ref{fig:reduction} gives
the histograms of the condition number reductions. We obtained these graphs by
sampling a hundred networks and computing the ratio of the condition number
before and after preconditioning. On the left we have the convex case, and on
the right the non-convex case. We clearly observe that the Jacobi and equilibration
method are closely matched for the convex case. However, in the non-convex case
equilibration significantly outperforms the other methods. Note that the poor
performance of the Gauss-Newton diagonal only means that its success in
optimization is not due to preconditioning. As we will see in Section
\ref{sec:results} these results extend to practical high-dimensional problems.

\section{Implementation}\label{sec:equilibration}

We propose to build a scalable algorithm for preconditioning neural networks
using equilibration. This method will estimate the same curvature information $\sqrt{\text{diag}({\bf H}^2)}$
with the unbiased estimator described in Equation \ref{eqn:equilibration}. It
is prohibitive to compute the full expectation at each learning step. Instead
we will simply update our running average at each learning step much like
RMSProp. The pseudo-code is given in Algorithm \ref{alg:equilibration}. The cost of this
is one product with the Hessian which is roughly
the cost of two additional gradient calculations and the cost of sampling a
vector from a random Gaussian. In practice we greatly amortize the cost by only
performing the update every 20 iterations. This brings the cost of equilibration
very close to that of regular SGD. The only added hyper-parameter is the damping
$\lambda$. We find that a good setting for that hyper-parameter is
$\lambda = 10^{-4}$ and it is robust over the tasks we considered.

\begin{figure}[t]
	\centering
    \begin{minipage}{1.0\textwidth}
    \begin{algorithm}[H]
   \caption{Equilibrated Gradient Descent}
   \label{alg:equilibration}
\begin{algorithmic}
        \REQUIRE Function $f(\theta)$ to minimize, learning rate $\epsilon$ and damping factor $\lambda$

		\STATE ${\bf D} \gets 0$

        \FOR{$i = k \to K$}
        \STATE ${\bf v} \sim \mathcal{N}(0, 1)$
        \STATE ${\bf D} \gets {\bf D} + ({\bf H}{\bf v})^2$
        \STATE $\theta \gets \theta - \epsilon \frac{\nabla f(\theta)}{\sqrt{{\bf D}/k} + \lambda}$
        \ENDFOR
    \end{algorithmic}
\end{algorithm}
\end{minipage}
\vskip -0.2in
\end{figure}

In the interest of comparison, we will evaluate SGD preconditioned with the
Jacobi preconditioner. This will allow us to verify the claims that
the equilibration preconditioner is better suited for non-convex problems.
\citet{bekas2007estimator} show that the diagonal of a matrix can be
recovered by the expression
\begin{align}
\text{diag}({\bf H}) = \mathrm{E}[{\bf v} \odot {\bf H}{\bf v}]
\end{align}
where ${\bf v}$ are random vectors with entries $\pm 1$ and $\odot$ is the
element-wise product. We use this estimator to precondition SGD in the same
fashion as that described in Algorithm \ref{alg:equilibration}. The variance
of this estimator for an element $i$ is $\sum_j H_{ji}^2 - H_{ii}^2$, while the method in
\citet{martens2012estimating} has $H_{ii}^2$. Therefore, the optimal method
depends on the situation. The computational
complexity is the same as ESGD.

\section{Experimental setup}\label{sec:setup}
We aim to confirm experimentally the theoretical results proposed in the
previous sections. Mainly, we want to verify that there is a significant
difference in performance between
the Jacobi preconditioner and equilibration methods for high-dimensional
non-convex problems.

In these experiments, we consider the challenging optimization benchmark of
training very deep neural networks. Following
\cite{martens2010hessian-small,sutskeverimportance,vinyals2011krylov}, we train
deep auto-encoders which have to reconstruct their input under the constraint
that one layer is very low-dimensional. This makes the reconstruction task
difficult because it requires the optimizer to finely tune the parameters. The
low-dimensional bottleneck layer learns a low-dimensional map similar in
principle to non-linear Principal Component Analysis. The networks have up to
11 layers of sigmoidal hidden units and have on the order of a million
parameters. We use the standard network architectures described in
\cite{martens2010hessian-small} for the MNIST and CURVES dataset. Both of
these datasets have 784 input dimensions and 60,000 and 20,000 examples respectively.

We tune the hyper-parameters of the optimization methods with random search. We
have sampled the learning rate from a logarithmic scale between $[0.1, 0.01]$
for stochastic gradient descent (SGD) and equilibrated SGD (ESGD). The learning
rate for RMSProp and the Jacobi preconditioner are sampled from
$[0.001, 0.0001]$. The damping factor $\lambda$ used
before dividing the gradient is taken from either $\{10^{-4},10^{-5},10^{-6}\}$
while the exponential decay rate of RMSProp is taken from either $\{0.9, 0.95\}$.
The networks are initialized using the sparse initialization described in
\cite{martens2010hessian-small}. We initialize 15 connections per neuron with
a zero mean unit variance Gaussian and the others are set to zero. The minibatch
size for all methods in 200. We do not
make use of momentum in these experiments in order to evaluate the strength
of each preconditioning method on its own. Similarly we do not use any
regularization because we are only concerned with optimization performance. For these reasons,
we report training error in our graphs. The networks and algorithms were implemented using Theano
\cite{Bastien-Theano-2012}, simplifying the use of the R-operator in
Jacobi and equilibrated SGD. All experiments were run on GPU's.

\section{Results}\label{sec:results}

\vspace*{-2mm}
\subsection{Comparison of preconditioned SGD methods}

\begin{figure*}[t]
\vskip -0.2in
\begin{center}
\subfigure[MNIST]{\includegraphics[width=0.45\linewidth]{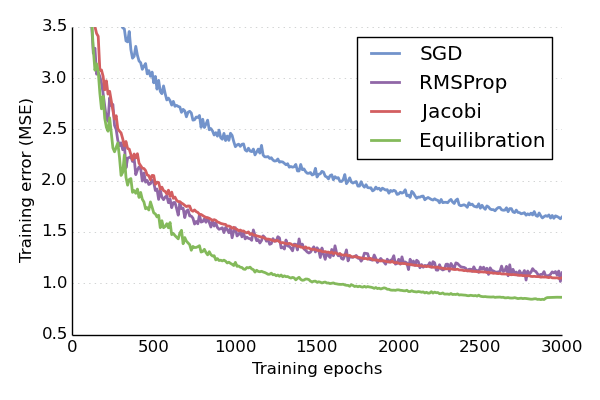}}
\subfigure[CURVES]{\includegraphics[width=0.45\linewidth]{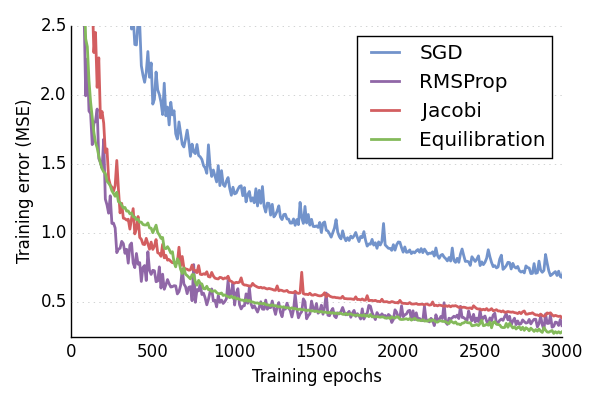}}
\vskip -0.1in
\caption{Learning Curves for deep auto-encoders on a) MNIST and b) CURVES comparing the
different preconditioned SGD methods. }
\label{fig:comparison}
\end{center}
\vskip -0.2in
\end{figure*}

We compare the different adaptive learning rates for training deep auto-encoders
in Figure \ref{fig:comparison}. We don't use momentum to better isolate
the performance of each method. We believe this is important because RMSProp has been
found not to mix well with momentum \citep{tieleman2012lecture}. Thus the results presented are not
state-of-the-art, but they do reach state of the art when momentum is used.

Our results on MNIST show that the proposed ESGD method significantly outperforms
both RMSProp and Jacobi SGD. The difference in performance becomes especially
notable after $250$ epochs. \cite{sutskeverimportance} reported a performance of 2.1 of training MSE
for SGD without momentum and we can see all adaptive learning rates improve on this result,
with equilibration reaching 0.86. We observe a convergence
speed that is approximately three times faster then our baseline SGD.
ESGD also performs best for CURVES, although the difference with
\mbox{RMSProp} and Jacobi SGD is not as significant as for MNIST. We postulated that
the smaller gap in performance might be due to the different preconditioners behaving
the same way on this dataset. We confirm this hypothesis in the next section.

\subsection{Measuring the similarity of the methods}\label{sec:rmspropbias}

\begin{figure*}[t]
\begin{center}
\subfigure[MNIST]{\includegraphics[width=0.45\linewidth]{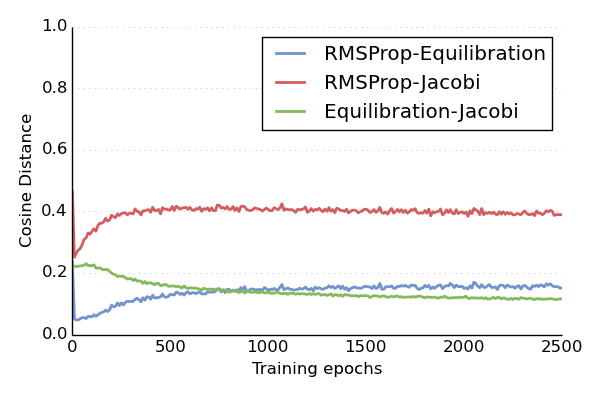}}
\subfigure[CURVES]{\includegraphics[width=0.45\linewidth]{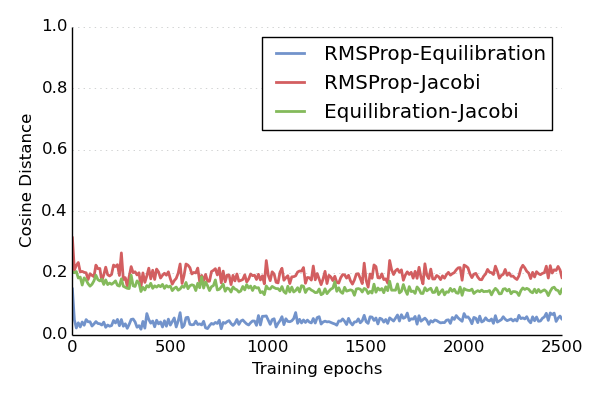}}
\vskip -0.1in
\caption{Cosine distance between the diagonals estimated by each method during the training of a deep auto-encoder trained on a) MNIST and b) CURVES. We can see that RMSProp estimates a quantity close to the equilibration matrix.}
\label{fig:bias}
\end{center}
\vskip -0.1in
\end{figure*}

We train deep autoencoders with RMSProp and measure every $10$ epochs the equilibration matrix $\mathbf{D}^{\text{E}} = \sqrt{\text{diag}(\mathbf{H}^2)}$ and Jacobi matrix $\mathbf{D}^{\text{J}} = \sqrt{\text{diag}(\mathbf{H})^2}$ using $100$ samples of the unbiased estimators described in Equations \ref{eqn:equilibration}, respectively. We then measure the pairwise differences between these quantities in terms of the cosine distance $\text{cosine}(u, v) = 1 - \frac{u \cdot v}{\|u\| \|v\|}$,
which measures the angle between two vectors and ignores their norms. Note further that the cosine distance ranges from 0 (zero degree angle) to 2 (180 degrees angle), and that the cosine distance between two preconditioners should not exceed one for preconditioned SGD to converge.

Figure \ref{fig:bias} shows the resulting cosine distances over training on MNIST and CURVES. For the latter dataset we observe that RMSProp remains remarkably close (around 0.05) to equilibration, while it is significantly different from Jacobi (in the order of 0.2). The same order of difference is observed when we compare equilibration and Jacobi, confirming the observations of Section \ref{sec:absolute} that both quantities are rather different in practice. For the MNIST dataset we see that RMSProp fairly well estimates $\sqrt{\text{diag}(\mathbf{H})^2}$ in the beginning of training, but then quickly diverges. After $1000$ epochs this difference has exceeded the difference between Jacobi and equilibration, and RMSProp no matches equilibration. Interestingly, at the same time that RMSProp starts diverging, we observe in Figure \ref{fig:comparison} that also the performance of the optimizer drops in comparison to ESGD. This may suggests that the success of RMSProp as a optimizer is tied to its similarity to the equilibration matrix.

\vspace*{-4mm}
\section{Conclusion}
We have studied diagonal preconditioners for saddle point problems i.e. indefinite matrices. We have shown by theoretical and empirical arguments that the equilibration preconditioner is comparatively better suited to this kind of problems than the Jacobi preconditioner. Using this insight, we have proposed a novel adaptive learning rate schedule for non-convex optimization problems, called ESGD, which empirically outperformed RMSProp on two competitive deep autoencoder benchmark. Interestingly, we have found that the update direction of RMSProp was in practice very similar to the equilibrated update direction, which might provide more insight into why RMSProp has been so successfull in training deep neural networks. More research is required to confirm these results. However, we hope that our findings will contribute to a better understanding of SGD's adaptive learning rate schedule for large scale, non-convex optimization problems.

%\section*{Acknowledgments}

\bibliography{paper,ml,aigaion}
\bibliographystyle{icml2015}

\end{document}